\documentclass{article}

\usepackage{arxiv}

\usepackage[utf8]{inputenc} 
\usepackage[T1]{fontenc}    
\usepackage{hyperref}       
\usepackage{url}            
\usepackage{booktabs}       
\usepackage{amsfonts}       
\usepackage{nicefrac}       
\usepackage{microtype}      
\usepackage{lipsum}		
\usepackage{graphicx}
\usepackage{natbib}
\usepackage{doi}

\usepackage{enumitem}
\usepackage{todonotes}
\usepackage{multirow}
\usepackage{multicol}
\usepackage{hyperref}
\usepackage{amsmath}
\usepackage{xspace}
\usepackage{mathtools}
\usepackage{array}
\usepackage[labelfont=bf]{caption}
\usepackage{subcaption}

\newcommand{\kg}{\ensuremath{\mathcal{G}}\xspace}

\newcommand{\query}{\ensuremath{\mathcal{Q}}\xspace}

\newcommand{\answer}{\ensuremath{\mathcal{A}}\xspace}

\newcommand{\prompt}{\ensuremath{\mathbf{t}}\xspace}

\title{Injecting Knowledge Graphs\\ into Large Language Models}

\author{\href{https://orcid.org/0000-0002-4670-8157}{\includegraphics[scale=0.06]{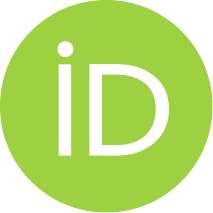}\hspace{1mm}Erica Coppolillo} \\
	University of Calabria\\
        ICAR-CNR\\
	\texttt{erica.coppolillo@unical.it}
    }

\date{}

\renewcommand{\shorttitle}{Let Your Graph Do the Reasoning}

\hypersetup{
pdftitle={A template for the arxiv style},
pdfsubject={q-bio.NC, q-bio.QM},
pdfauthor={David S.~Hippocampus, Elias D.~Striatum},
pdfkeywords={First keyword, Second keyword, More},
}

\begin{document}

\maketitle

\begin{abstract}
Integrating structured knowledge from Knowledge Graphs (KGs) into Large Language Models (LLMs) remains a key challenge for symbolic reasoning. Existing methods mainly rely on prompt engineering or fine-tuning, which lose structural fidelity or incur high computational costs. Building on recent encoding techniques which integrate graph embeddings within the LLM input as tokens, we extend this paradigm to the KG domain by leveraging Knowledge Graph Embedding (KGE) models, thus enabling graph-aware reasoning. Our approach is model-agnostic, resource-efficient, and compatible with any LLMs. Extensive experimentation on synthetic and real-world datasets shows that our method improves reasoning performance over established baselines, further achieving the best trade-off in terms of accuracy and efficiency against state-of-the-art LLMs. 
\end{abstract}





\section{Introduction}

Large Language Models (LLMs) have demonstrated impressive performance across a range of natural language tasks,  including machine translation~\citep{gain2025bridginglinguisticdividesurvey}, question answering~\citep{yue2025survey}, summarization~\citep{10.1145/3731445}, and dialogue generation~\citep{chen-etal-2024-recent}. However, they suffer from limitations such as hallucinations, lack of explicit memory, and difficulty in handling \textit{structured knowledge}, such as the relational information encoded in Knowledge Graphs (KGs)~\citep{ji2021survey}. 
Knowledge Graphs represent facts as networks of entities and relations, offering a form of symbolic knowledge that enables logical inference and precise factual retrieval.

Integrating the rich structured context of KGs into the generative reasoning abilities of LLMs is a promising direction,
yet achieving effective integration remains challenging.
Indeed, most existing approaches rely on \textit{prompting} strategies~\citep{biran2024hopping, chen2024new, wu2024thinking}, where graph facts are serialized into text and used as context. While simple to implement, these methods often lose relational structure and require careful template engineering. Other strategies~\citep{zhao2023kgcot} use dedicated reasoning modules to traverse KGs and then guide the LLM through chain-of-thought prompting. Fine-tuning-based methods~\citep{sun2023llmkg} adapt the LLM on graph-specific tasks but incur high computational costs and risk overfitting.

We propose a novel alternative: \textbf{injecting structured knowledge graph representations directly into a frozen LLM}. Our approach builds on recent work~\citep{perozzi2024letgraphtalkingencoding} which introduces GraphToken, a token-based encoding scheme for generic graphs. We extend this idea to the \textit{knowledge graph} setting by leveraging Knowledge Graph Embedding (KGE) models to encode entities and relations as vectors. These vectors are transformed into structured tokens and injected into the LLM input alongside the natural language query.

Importantly, our method does not require any LLM fine-tuning or prompt engineering, is model-agnostic, and resource-efficient. The structured tokens are generated at inference time using trained KGE models and do not involve any weight updates to the LLM. As such, the technique can be integrated with any LLM that accepts input sequences.

We evaluate our method on both synthetic tasks and real-world datasets from the graph learning literature, providing a diverse testbed for our experimental evaluation.

\noindent{Our main contributions are the following:}
\begin{itemize}[leftmargin=*]
    \item We extend the GraphToken framework~\citep{perozzi2024letgraphtalkingencoding} to the knowledge graph domain, enabling structured injection of KG information into LLMs.
    \item Our method eliminates the need for LLM training or prompt engineering, operating directly with frozen LLMs. The only trainable component is the KGE model, ensuring efficiency and resource savings.
    \item We demonstrate the effectiveness and generalizability of the framework across both synthetic and real-world KG datasets and against several competitors, using a lightweight LLM as the underlying backbone.
\end{itemize}
\vspace{0.1cm}
The remainder of the paper is organized as follows: Section~\ref{sec:related} reviews related work on enhancing LLM reasoning over graphs. Section~\ref{sec:methodology} introduces our proposed methodology, followed by experimental setup in Section~\ref{sec:experiments}. The results are provided in Section~\ref{sec:results}. We conclude in Section~\ref{sec:conclusions}, where we discuss limitations and future directions.

\section{Related Work}
\label{sec:related}

\paragraph{KGQA} Question Answering over Knowledge Graphs (KGQA) has evolved significantly, transitioning from early rule-based systems to sophisticated models integrating deep learning and large language models (LLMs). Initial approaches focused on semantic parsing techniques to convert natural language questions into structured queries over knowledge graphs \citep{berant2013semantic, yih2015semantic}.

With the rise of neural networks, embedding-based methods gained prominence. Bordes et al.~\citep{bordes2014qaembedding} introduced a model that learns low-dimensional embeddings for entities and relations, facilitating question answering through vector space representations. Subsequent works like GraftNet \citep{sun2018graftnet} and PullNet \citep{sun2019pullnet} employed graph neural networks to capture multi-hop reasoning paths within knowledge graphs.

\paragraph{LLMs for KG Reasoning}  Recent advancements have seen the integration of LLMs with knowledge graphs to enhance reasoning capabilities. For instance, QA-GNN \citep{yasunaga2021qagnn} combines language models with graph neural networks to jointly reason over textual and structured data. Similarly, approaches like R3 \citep{toroghi2024right} aim to ground LLM outputs in knowledge graphs, ensuring verifiable and factually accurate answers.
Indeed, a common strategy is to \textit{prompt} the LLM with KG-derived context. For example, \citet{huang2023llmkg} showed that including relevant triples in prompts improves factual QA on long-tail queries. Other works use structured prompts or reasoning traces to further guide the model. \citet{zhao2023kgcot} proposed KG-CoT, a method that traverses the KG to generate a step-by-step reasoning path, which is then provided to the LLM as a chain-of-thought prompt. Such methods have demonstrated improved performance on multi-hop and open-domain KGQA tasks without fine-tuning the LLM.

\paragraph{Graph Encodings for LLMs} Beyond prompting, another line of work explores how to encode graph structure into LLM-consumable formats. \citet{fatemi2023talklikegraphencoding} and \citet{liu2024graphqa} investigated how different graph-to-text serializations affect reasoning performance, revealing that appropriate encodings can yield large accuracy gains. Other methods, such as soft prompts~\citep{liu2023softgraphprompt} represent the graph in vector form and append this to the LLM input. While effective, these methods often require either extensive prompt design or additional model training.
\newline

Our work departs from the above paradigms by injecting knowledge graphs into an LLM through structured embeddings, without any LLM training or prompt engineering. Extending the work of~\citet{perozzi2024letgraphtalkingencoding} from generic, unlabeled graphs to KGs, we leverage Knowledge Graph Embedding (KGE) models to convert entities and relations into vector representations, which are then supplied to a frozen LLM. To the best of our knowledge, this is the first method to enable KG reasoning in LLMs using structured injection from KGE models. 

\section{Methodology}
\label{sec:methodology}
\paragraph{Reasoning Tasks}
\begin{table*}[!ht]
    \centering
    \caption{Template of the considered queries for the identification task, embracing 0-, 1-, and 2-hop neighborhood questions, and   Single, Double, and Triplet Label type.}
    \label{tab:query-examples}
    \resizebox{\linewidth}{!}{
    \begin{tabular}{ccl}
         \toprule
         {\textbf{Hop}} & {\textbf{Type}} & \textbf{Template Query} \\
         \midrule
         0 & SL & 
         {Which nodes are \texttt{[LABEL]}s in the graph?} \\
         \cmidrule(lr){2-3}
         \multirow{2}{*}{1} & SL & 
         {Which nodes have a \texttt{[LABEL]} as neighbor?} \\
                            & DL & 
                            {Which nodes are \texttt{[LABEL1]}s and have a \texttt{[LABEL2]} as neighbor?}\\
         \cmidrule(lr){2-3}
         \multirow{3}{*}{2} & SL & 
                             {Which nodes have a neighbor who \texttt{[PROPERTY]}s \texttt{[LABEL]}s?} \\
                            
                            & DL & 
                            {Which nodes are \texttt{[LABEL1]}s and have a neighbor who \texttt{[PROPERTY]}s \texttt{[LABEL2]}s?} \\
         & TL & 
         {Which nodes are \texttt{[LABEL1]}s and have a \texttt{[LABEL2]} as neighbor who \texttt{[PROPERTY]}s \texttt{[LABEL3]}s?} \\
        \bottomrule
    \end{tabular}
    }
\end{table*}
To evaluate the reasoning capabilities of the model, we consider the following node reasoning tasks: {Existence} (\textbf{E}), {Counting} (\textbf{C}), and Identification ({\textbf{I}}), progressively increasing the complexity of the query by supporting \textbf{0-Hop}, \textbf{1-Hop}, and \textbf{2-Hop} neighborhood. Specifically, we examine the following query categories:
\begin{itemize}[leftmargin=*]
    \item 0-Hop, Single Label (\textbf{SL}): query involves labeled nodes.
    \item 1-Hop, Single Label: query involves nodes having labeled neighbors. 
    \item 1-Hop, Double Label (\textbf{DL}): query involves labeled nodes having labeled neighbors.
    \item 2-Hop, Single Label: query involves nodes having neighbors with a property regarding other labeled nodes.
    \item 2-Hop, Double Label: query involves labeled nodes having neighbors with a property regarding other labeled nodes.
    \item 2-Hop, Triple Label (\textbf{TL}): query involves labeled nodes having labeled neighbors with a property regarding other labeled nodes.
\end{itemize}

To ensure comprehension, Table~\ref{tab:query-examples} provides the query template for each considered category on the identification task. The instantiation is equivalent to the existence and counting tasks, trivially reformulating the query with ``Are there~\dots?'' and ``How many~\dots?'', respectively, instead of ``Which~\dots?''.
\paragraph{GraphToken} We here recall some key concepts of GraphToken, the encoding method introduced by~\citet{perozzi2024letgraphtalkingencoding}, to encode graph structures into continuous vector representations, which we extend to knowledge graphs. The GraphToken framework mainly consists of a Graph Encoder, which utilizes a Graph Neural Network (GNN)~\citep{ZHOU202057} to process the input graph, capturing its structural information. Node features are derived using positional encodings, such as Laplacian eigenvectors or learned embeddings. For further technical details, please refer to the main paper.

\paragraph{KG Embedding} To compute the latent representation of the KG, we rely on  
Knowledge Graph Embedding (KGE) models~\citep{kge-models}, which aim to represent the entities and relations in a knowledge graph as low-dimensional vectors in a continuous vector space. 

Given a set of entities $\mathcal{E}$ and relations $\mathcal{R}$, a knowledge graph \kg is typically represented as a set of triples \( \mathcal{T} = \{(h, r, t)\} \), where \( h \) and \( t \) are the head and tail entities, and \( r \) is the relation.

Each entity \( h, t \in \mathcal{E} \) and each relation \( r \in \mathcal{R} \) is mapped to an embedding vector:
\[
\mathbf{e}_h, \mathbf{e}_t, \mathbf{e}_r \in \mathbb{R}^d
\]

A scoring function \( f(h, r, t) \) is defined to measure the plausibility of a triple, where the form of \( f \) depends on the specific KGE model.

To train the model, a margin-based ranking loss is typically used~\citep{rankingloss1, ranking2, ranking3, ranking4, ranking5}:

\begin{equation}
\label{eq:ranking-loss}
\begin{split}
\mathcal{L} = & \sum_{(h, r, t) \in \mathcal{T}} \sum_{(h', r, t') \in \mathcal{T}'} \max\big(0, \gamma + f(h', r, t') - f(h, r, t)\big)
\end{split}
\end{equation}


where \( \gamma > 0 \) is the margin, and \( \mathcal{T}' \) is the set of negative samples generated by corrupting either the head or tail entity.

The model parameters (entity and relation embeddings) are trained using stochastic gradient descent~\citep{ruder2017overviewgradientdescentoptimization} or variants (e.g., Adam~\citep{kingma2017adammethodstochasticoptimization}). Embeddings are often normalized during training to enforce constraints (e.g., unit norm):

\begin{equation}
\mathbf{e}_h \leftarrow \frac{\mathbf{e}_h}{\|\mathbf{e}_h\|}, \quad \mathbf{e}_t \leftarrow \frac{\mathbf{e}_t}{\|\mathbf{e}_t\|}, \quad \mathbf{e}_r \leftarrow \frac{\mathbf{e}_r}{\|\mathbf{e}_r\|}
\end{equation}

To our purpose, instead of aggregating the scalar scores as in Equation~\ref{eq:ranking-loss}, we rewrite the aggregation function to preserve per-dimension contributions. More formally, let \(\mathbf{x} \in \mathbb{R}^{n \times d}\) be a matrix representing \(n\) KG triplets with \(d\) features each. We define the new scoring vector $\mathbf{s} \in \mathbb{R}^d$ via the aggregation function $f'(\mathbf{x})$ as follows:

\begin{align}
\mathbf{s} = f'(\mathbf{x}) &= \begin{bmatrix}
f(\mathbf{x}_{:,1}), f(\mathbf{x}_{:,2}),~\cdots, f(\mathbf{x}_{:,d})
\end{bmatrix}.
\end{align}

Intuitively, in this formulation, while $f(\mathbf{x})$ performs row-wise aggregation across the feature dimension, producing a scalar per sample,  $f'(\mathbf{x})$ performs column-wise aggregation across the sample dimension, yielding a single representative feature vector. This final vector can be interpreted as a dimension-wise decomposition of the graph's triples plausibility~\citep{Holtz2011}.

Finally, to make this representation compatible with the token embedding space of the LLM, we apply a dense layer, i.e., a trainable linear projection:

\begin{equation}
\label{eq:kg-embedding}
    \mathbf{g} = \mathbf{W} \mathbf{s} + \mathbf{b}, \quad \mathbf{g} \in \mathbb{R}^{d_{\text{LLM}}}
\end{equation}

where \( \mathbf{W} \in \mathbb{R}^{d_{\text{LLM}} \times d} \) and \( \mathbf{b} \in \mathbb{R}^{d_{\text{LLM}}} \) are learnable parameters, and \( d_{\text{LLM}} \) is the dimensionality of the (frozen) LLM token embeddings.

The resulting vector \( \mathbf{g} \) thus constitutes the final latent representation of \kg.

\paragraph{Training Process}
Similarly to the original work~\citep{perozzi2024letgraphtalkingencoding}, the training dataset comprises pairs $(\query, \answer)$, where \query includes the textual description of a knowledge graph \kg with the corresponding reasoning question; and \answer represents the ground truth answer.

During the forward pass, the final input encoding \prompt is obtained by concatenating the tokenized representation of the query \query with the embedding of \kg:

\[
\prompt = \mathbf{q} \mathbin\Vert \mathbf{g}
\]

where $\mathbf{q}$ denotes the tokenized form of the query, and $\mathbf{g}$ is computed following Equation~\ref{eq:kg-embedding}.

The objective is to minimize the negative log-likelihood loss~\citep{nll} of the LLM's output conditioned on the augmented input \prompt:

\begin{equation}
\label{eq:nll-loss}
\mathcal{L}(\answer \mid \query) = -\log \textrm{P}_{\text{LLM}}(\answer \mid \prompt)
\end{equation}

\begin{figure*}
    \centering
    \includegraphics[width=\linewidth]{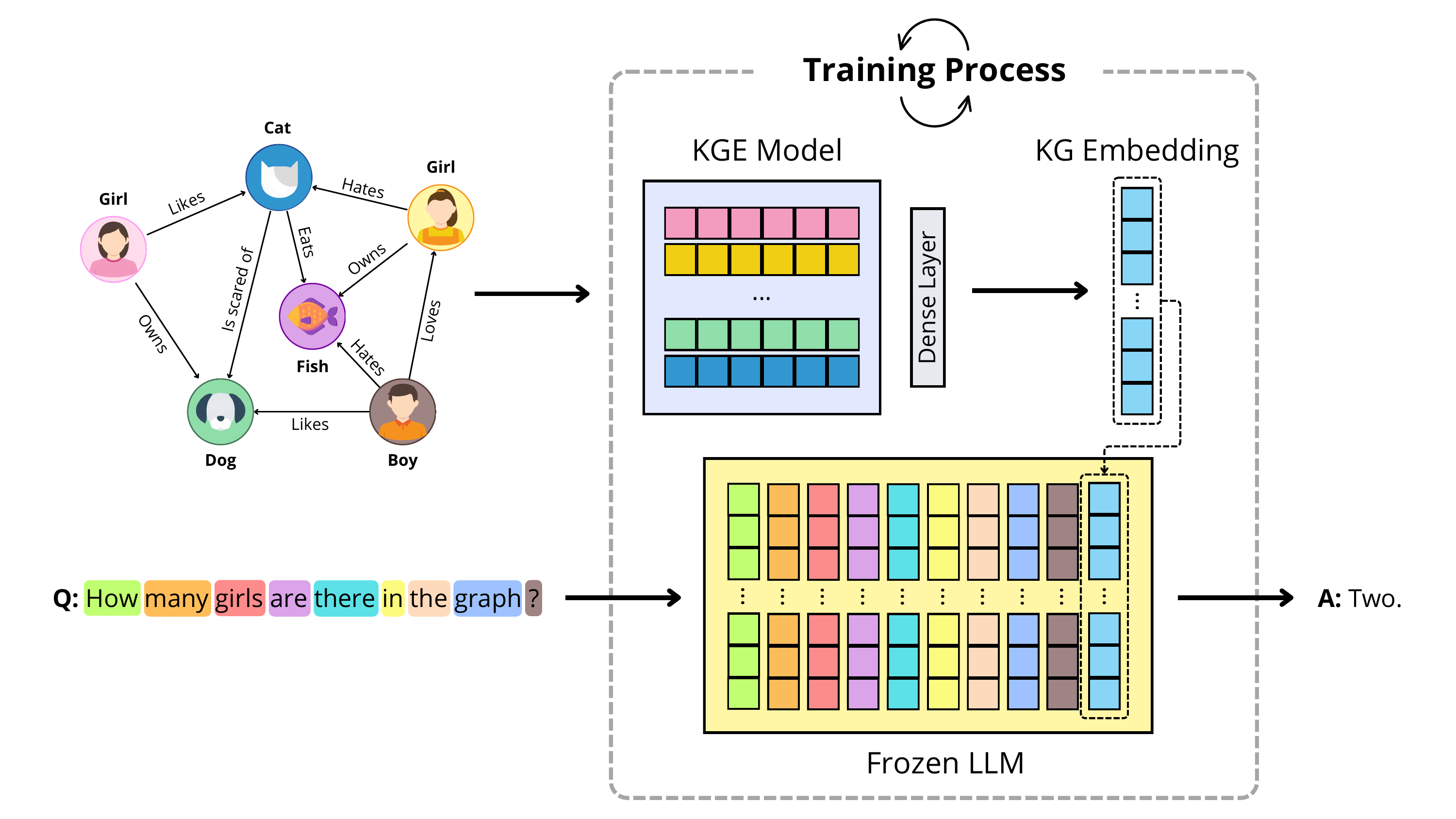}
    \caption{Visual representation of the framework. During training, the KG embedding is concatenated to the token latent vectors of the frozen LLM, and the answer produced by the LLM is optimized.}
    \label{fig:framework}
\end{figure*}

During training, the parameters of the LLM are always kept \textbf{frozen}. Only the parameters of the underlying KGE model are updated, by back-propagating to minimize $\mathcal{L}(\answer \mid \query)$. A sketch of the overall framework is depicted in Figure~\ref{fig:framework}.

\section{Experimental Setup}
\label{sec:experiments}
The experimental evaluation of our methodology aims at answering the following research questions:
\begin{itemize}[leftmargin=1.05cm]
    \item[\textbf{RQ1}:] How does the proposed strategy compare on reasoning tasks against state-of-the-art methods and baselines?
    \item[\textbf{RQ2}:] How do the results vary by changing the underlying KGE model?
    \item[\textbf{RQ3}:] Is the final graph embedding effectively tailored to the given reasoning task?
\end{itemize}

\begin{table}[!ht]
    \centering
    
    \caption{Knowledge graphs statistics from both synthetic and real-world datasets.}
    \label{tab:datasets-statistics}
    \vspace{0.2cm}
    \resizebox{0.5\columnwidth}{!}{
    \begin{tabular}{cccc}
    \toprule
         \textbf{Dataset} & \textbf{\#Graphs} & \textbf{Avg. \#Nodes} & \textbf{Avg. \#Edges}  \\
    \midrule
        Synthetic & 600 & 11.91 & 74.46 \\
        AIDS & 2,000 & 15.59 & 32.39 \\
        MUTAG & 187 & 18.03 & 39.8 \\
        AQSOL & 9,833 & 17.6 & 35.8 \\
    \bottomrule
    \end{tabular}
    }  
\end{table}
\paragraph{Datasets} To assess the validity of our method, we perform the experiments by relying on both synthetic and real-world datasets, described as follows: 
\begin{itemize}[leftmargin=*]
    \item \textbf{Synthetic}: We synthetically generate undirected, complete graphs of varying sizes, labeling nodes and edges with randomly chosen entities and relations, respectively.
    \item \textbf{AIDS}\footnote{\url{https://huggingface.co/datasets/graphs-datasets/AIDS}}~\citep{aids1, aids2}: The AIDS dataset consists of molecular graphs representing chemical compounds checked for evidence of anti-HIV activity. Nodes and edges represent chemical bonds and are provided with the corresponding features. 
    \item \textbf{MUTAG}\footnote{\url{https://huggingface.co/datasets/graphs-datasets/MUTAG}}~\citep{mutag}: MUTAG is a dataset of nitroaromatic compounds, represented as graphs. Nodes are atoms (with labels such as C, N, O), and edges are chemical bonds. Each compound is labeled based on its mutagenic effect on a bacterium (Salmonella typhimurium).
    \item \textbf{AQSOL}\footnote{\url{https://huggingface.co/datasets/graphs-datasets/AQSOL}}~\citep{aqsol}: AQSOL is a dataset of small molecules with their aqueous solubility values. Molecules are typically represented using SMILES strings and can be converted into molecular graphs. 
\end{itemize}

Table~\ref{tab:datasets-statistics} provides some statistics, including the number of graphs within each dataset, and the average number of nodes and edges.

\paragraph{KGE Models}
In our experiments, we implement and adapt the following KGE models to compute the knowledge graph embedding:

\begin{itemize}[leftmargin=*]
    \item \textbf{TransE} \citep{transe}: Models relations as translations in the embedding space. For a triple $(h, r, t)$, it enforces $\mathbf{e}_h + \mathbf{e}_r \approx \mathbf{e}_t$. The score function is $f(h, r, t) = -\|\mathbf{e}_h + \mathbf{e}_r - \mathbf{e}_t\|_p$, where $p$ refers to the norm type used to compute the distance. 

    \item \textbf{DistMult} \citep{distmult}: Represents entities and relations as real-valued vectors and models interactions via element-wise multiplication. The score function is $f(h, r, t) = \langle \mathbf{e}_h, \mathbf{e}_r, \mathbf{e}_t \rangle = \sum_i e_h^i e_r^i e_t^i$. 

    \item \textbf{ComplEx} \citep{complex}: Extends DistMult to complex-valued embeddings, enabling the modeling of asymmetric relations. It uses the score function $f(h, r, t) = \Re(\langle \mathbf{e}_h, \mathbf{e}_r, \overline{\mathbf{e}}_t \rangle)$, where $\overline{\mathbf{e}}_t$ is the complex conjugate of $\mathbf{e}_t$, and $\Re$ denotes the real part.
    \item \textbf{RotatE} \citep{rotate}: Embeds entities in the complex space and represents relations as rotations. For each triple, it models $\mathbf{t} \approx \mathbf{h} \circ \mathbf{r}$, where $\circ$ is the element-wise complex multiplication and $\mathbf{r}$ lies on the unit circle. The score function is $f(h, r, t) = -\|\mathbf{e}_h \circ \mathbf{e}_r - \mathbf{e}_t\|$. 
\end{itemize}

\paragraph{LLM} For the experiments, we adopt the 2B, fine-tuned version of Gemma~\citep{gemma}, a lightweight model based on Gemini\footnote{\url{https://deepmind.google/technologies/gemini/}}, the LLMs family released by Google. However, we remark that the framework is independent from the underlying model since \textbf{any} LLM can be used.   

\paragraph{Competitors}
We compare our method against a comprehensive set of established baselines:

\begin{itemize}[leftmargin=*]
    \item \textbf{Zero-Shot}: The model receives only a task description and is immediately asked to generate the desired output, without any examples or demonstrations.

    \item \textbf{Few-Shot}~\citep{few-shot}: The model is provided with a few example input-output pairs, embedded directly in the prompt. This enables the model to adapt during inference, without conventional training. In our experiments, the number of provided examples is equal to 2.

    \item \textbf{CoT} (\textbf{C}hain-\textbf{o}f-\textbf{T}hought)~\citep{cot}: This prompting technique includes examples with step-by-step reasoning, encouraging the model to generate intermediate steps when solving new tasks.

    \item \textbf{Zero-CoT}~\citep{zero-cot}: A variant of CoT prompting that requires no examples. Instead, it elicits reasoning by using a simple trigger phrase like ``Let’s think step by step''.

    \item \textbf{CoT-BaG}~\citep{cot-bag}: An extension of CoT integrating \textbf{B}uilding-\textbf{a}-\textbf{G}raph prompting, which adds a statement to guide the model’s reasoning on graph-related tasks.

    \item \textbf{Prompt Tuning}~\citep{prompt-tuning}: A parameter-efficient fine-tuning technique where a small set of learnable embeddings (called \textit{soft prompt}) is prepended to the model’s input. Instead of updating the full model, only these prompt parameters are trained, allowing the LLM to adapt to new tasks with minimal computational cost.
\end{itemize}

Additionally, we test our reasoning framework against the following state-of-the-art LLMs, to evaluate their trade-off between accuracy and efficiency: \textbf{GPT-4o}~\citep{gpt4o}, a powerful model produced by OpenAI in 2024, capable of processing and generating text, audio, and images; \textbf{o4-mini}~\citep{o4mini}, a compact, cost-efficient reasoning model released by OpenAI in 2025, and optimized for tasks in mathematics, coding, and visual understanding; and \textbf{DeepSeek-R1}~\citep{deepseek2025r1}, a large-scale reasoning model developed by DeepSeek, trained via reinforcement learning for tasks in mathematics, coding, and reasoning.

\begin{table*}[!ht]
        \centering
         \caption{Test accuracy over the considered datasets and reasoning tasks, against the selected established baselines. Best scores are highlighted in bold. Percentage improvement is computed with respect to the second-best score.}
        \label{tab:unique-results}
        \resizebox{\linewidth}{!}{
        \begin{tabular}{w{c}{0.2cm}cccc|cccccc|ccccccccc}
        \toprule
        {} & {} & \multicolumn{3}{c}{\textbf{0-Hop}} & \multicolumn{6}{c}{\textbf{1-Hop}} & \multicolumn{9}{c}{\textbf{2-Hop}} \\
        \cmidrule(lr){3-5}
        \cmidrule(lr){6-11}
        \cmidrule(lr){12-20}
        {\textbf{Data}} & \textbf{Method} & \textbf{E-SL} & \textbf{C-SL} & \textbf{I-SL} & \textbf{E-SL} & \textbf{C-SL} & \textbf{I-SL} & \textbf{E-DL} & \textbf{C-DL} & \textbf{I-DL} & \textbf{E-SL} & \textbf{C-SL} & \textbf{I-SL} & \textbf{E-DL} & \textbf{C-DL} & \textbf{I-DL} & \textbf{E-TL} & \textbf{C-TL} & \textbf{I-TL} \\
        \midrule
        \multirow{8}{*}{\centering{\rotatebox[origin=c]{90}{\textbf{Synthetic}}}} & Zero-Shot & 0.43 & 0.05 & 0.0 & 0.43 & 0.05 & 0.05 & 0.0 & 0.07 & 0.0 & 0.68 & 0.05 & 0.05 & 0.02 & 0.1 & 0.0 & 0.03 & 0.2 & 0.05\\
& Few-Shot & 0.93 & 0.22 & 0.07 & 0.77 & 0.12 & 0.18 & 0.55 & 0.15 & 0.02 & 0.52 & 0.12 & 0.13 & 0.48 & 0.12 & 0.1 & 0.35 & 0.12 & 0.02\\
& CoT & 0.9 & 0.13 & 0.07 & 0.65 & 0.12 & 0.17 & 0.52 & 0.13 & 0.02 & 0.45 & 0.08 & 0.12 & 0.43 & 0.15 & 0.08 & 0.48 & 0.22 & 0.03\\
& Zero-CoT & 0.0 & 0.05 & 0.0 & 0.35 & 0.0 & 0.0 & 0.45 & 0.0 & 0.0 & 0.83 & 0.0 & 0.0 & 0.27 & 0.08 & 0.0 & 0.58 & 0.0 & 0.05\\
& CoT-BaG & 0.78 & 0.13 & 0.07 & 0.65 & 0.12 & 0.17 & 0.52 & 0.13 & 0.02 & 0.57 & 0.12 & 0.13 & 0.53 & 0.12 & 0.1 & 0.57 & 0.25 & 0.03\\
& Prompt Tuning & 0.77 & 0.17 & 0.03 & 0.52 & 0.15 & 0.35 & 0.3 & 0.1 & 0.0 & 0.28 & 0.02 & 0.12 & 0.0 & 0.07 & 0.03 & 0.03 & 0.12 & 0.02\\
\cmidrule(lr){2-20}
 & \textbf{Ours} & \textbf{1.0} & \textbf{0.7} & \textbf{0.08} & \textbf{1.0} & \textbf{0.88} & \textbf{0.93} & \textbf{1.0} & \textbf{0.63} & \textbf{0.13} & \textbf{1.0} & \textbf{0.88} & \textbf{0.93} & \textbf{0.97} & \textbf{0.67} & \textbf{0.13} & \textbf{0.95} & \textbf{0.7} & \textbf{0.18}\\
\midrule
\multirow{8}{*}{\centering{\rotatebox[origin=c]{90}{\textbf{AIDS}}}} & Zero-Shot & 0.45 & 0.08 & 0.01 & 0.45 & 0.12 & 0.07 & 0.0 & 0.28 & 0.04 & 0.71 & 0.12 & 0.07 & 0.0 & 0.07 & 0.0 & 0.08 & 0.45 & 0.18\\
& Few-Shot & 0.93 & 0.26 & 0.02 & 0.71 & 0.16 & 0.05 & 0.69 & 0.36 & 0.01 & 0.72 & 0.17 & 0.04 & 0.71 & 0.14 & 0.16 & 0.62 & 0.3 & 0.0\\
& CoT & 0.89 & 0.22 & 0.02 & 0.7 & 0.12 & 0.09 & 0.7 & 0.3 & 0.01 & 0.74 & 0.17 & 0.04 & 0.72 & 0.14 & 0.15 & 0.7 & 0.36 & 0.0\\
& Zero-CoT & 0.0 & 0.08 & 0.0 & 0.34 & 0.02 & 0.02 & 0.46 & 0.0 & 0.04 & 0.68 & 0.02 & 0.02 & 0.11 & 0.1 & 0.02 & 0.19 & 0.0 & 0.18\\
& CoT-BaG & 0.84 & 0.22 & 0.02 & 0.7 & 0.12 & 0.09 & 0.7 & 0.3 & 0.01 & 0.72 & 0.18 & 0.04 & 0.74 & 0.13 & 0.16 & 0.68 & 0.38 & 0.0\\
& Prompt Tuning & 0.07 & 0.08 & 0.0 & 0.67 & 0.06 & 0.04 & 0.14 & 0.08 & 0.02 & 0.59 & 0.08 & 0.12 & 0.1 & 0.09 & 0.14 & 0.49 & 0.1 & 0.02\\
\cmidrule(lr){2-20}
 & \textbf{Ours} & \textbf{1.0} & \textbf{0.4} & \textbf{0.06} & \textbf{1.0} & \textbf{0.28} & \textbf{0.18} & \textbf{0.95} & \textbf{0.62} & \textbf{0.43} & \textbf{0.97} & \textbf{0.25} & \textbf{0.18} & \textbf{0.9} & \textbf{0.39} & \textbf{0.18} & \textbf{0.95} & \textbf{0.64} & \textbf{0.46}\\
\midrule
\multirow{8}{*}{\centering{\rotatebox[origin=c]{90}{\textbf{MUTAG}}}} & Zero-Shot & 0.37 & 0.0 & 0.0 & 0.37 & 0.0 & 0.05 & 0.0 & 0.16 & 0.0 & 0.32 & 0.0 & 0.05 & 0.0 & 0.05 & 0.0 & 0.0 & 0.0 & 0.0\\
& Few-Shot & 0.89 & 0.11 & 0.05 & 0.63 & 0.0 & 0.0 & 0.53 & 0.16 & 0.05 & 0.63 & 0.11 & 0.0 & 0.63 & 0.16 & 0.16 & 0.63 & 0.16 & 0.05\\
& CoT & 0.95 & 0.26 & 0.05 & 0.47 & 0.0 & 0.0 & 0.63 & 0.32 & 0.11 & 0.63 & 0.16 & 0.0 & 0.63 & 0.26 & 0.05 & 0.68 & 0.16 & 0.11\\
& Zero-CoT & 0.0 & 0.05 & 0.0 & 0.42 & 0.0 & 0.0 & 0.16 & 0.0 & 0.0 & 0.32 & 0.0 & 0.0 & 0.0 & 0.0 & 0.0 & 0.32 & 0.0 & 0.0\\
& CoT-BaG & 0.89 & 0.32 & 0.05 & 0.47 & 0.0 & 0.0 & 0.63 & 0.32 & 0.11 & 0.63 & 0.16 & 0.0 & 0.58 & 0.26 & 0.05 & 0.68 & 0.16 & 0.11\\
& Prompt Tuning & 0.89 & 0.0 & 0.0 & 0.47 & 0.0 & 0.0 & 0.0 & 0.0 & 0.0 & 0.0 & 0.05 & 0.0 & 0.58 & 0.0 & 0.0 & 0.05 & 0.0 & 0.0\\
\cmidrule(lr){2-20}
 & \textbf{Ours} & \textbf{1.0} & \textbf{0.74} & \textbf{0.58} & \textbf{1.0} & \textbf{0.58} & \textbf{0.47} & \textbf{0.89} & \textbf{0.84} & \textbf{0.68} & \textbf{1.0} & \textbf{0.58} & \textbf{0.47} & \textbf{0.89} & \textbf{0.95} & \textbf{0.68} & \textbf{1.0} & \textbf{0.95} & \textbf{0.63}\\
\midrule
\multirow{8}{*}{\centering{\rotatebox[origin=c]{90}{\textbf{AQSOL}}}} & Zero-Shot & 0.53 & 0.06 & 0.01 & 0.51 & 0.15 & 0.02 & 0.03 & 0.33 & 0.0 & 0.73 & 0.15 & 0.02 & 0.17 & 0.07 & 0.0 & 0.08 & 0.37 & 0.06\\
& Few-Shot & 0.92 & 0.18 & 0.05 & 0.74 & 0.16 & 0.05 & 0.69 & 0.28 & 0.01 & 0.66 & 0.15 & 0.05 & 0.69 & 0.16 & 0.1 & 0.58 & 0.3 & 0.02\\
& CoT & 0.86 & 0.15 & 0.04 & 0.73 & 0.15 & 0.04 & 0.67 & 0.28 & 0.01 & 0.67 & 0.16 & 0.05 & 0.71 & 0.15 & 0.1 & 0.71 & 0.32 & 0.01\\
& Zero-CoT & 0.0 & 0.05 & 0.0 & 0.01 & 0.02 & 0.12 & 0.74 & 0.0 & 0.01 & 0.51 & 0.02 & 0.12 & 0.48 & 0.02 & 0.01 & 0.22 & 0.0 & 0.12\\
& CoT-BaG & 0.75 & 0.17 & 0.05 & 0.73 & 0.15 & 0.04 & 0.67 & 0.28 & 0.01 & 0.69 & 0.14 & 0.05 & 0.73 & 0.14 & 0.1 & 0.72 & 0.34 & 0.03\\
& Prompt Tuning & 0.69 & 0.1 & 0.02 & 0.04 & 0.01 & 0.01 & 0.52 & 0.06 & 0.01 & 0.31 & 0.08 & 0.01 & 0.01 & 0.06 & 0.06 & 0.34 & 0.04 & 0.01 \\
\cmidrule(lr){2-20}
 & \textbf{Ours} & \textbf{1.0} & \textbf{0.38} & \textbf{0.06} & \textbf{1.0} & \textbf{0.27} & \textbf{0.22} & \textbf{0.97} & \textbf{0.57} & \textbf{0.4} & \textbf{0.98} & \textbf{0.27} & \textbf{0.22} & \textbf{0.96} & \textbf{0.32} & \textbf{0.17} & \textbf{0.96} & \textbf{0.58} & \textbf{0.41}\\
    \bottomrule
        \end{tabular}
        }
       
    \end{table*}

\paragraph{Setting}
For the experiments, we split the dataset into training, validation and test sets, using a $80/10/10$ ratio. We trained the models by performing early stopping on the validation set~\citep{early-stopping}. To optimize Equation~\ref{eq:nll-loss}, we adopt a Lion optimizer~\citep{chen2023symbolic}, with a learning rate equal to $10e\rm{-}3$. Regarding the configuration of the KGE models, we set the hidden dimension to 64. 

\begin{figure}[!ht]
    \centering
    \begin{minipage}{0.7\linewidth}
        \includegraphics[width=\linewidth]{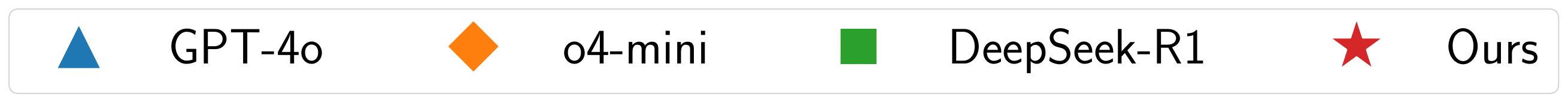}
    \end{minipage}
    \includegraphics[width=0.5\linewidth]{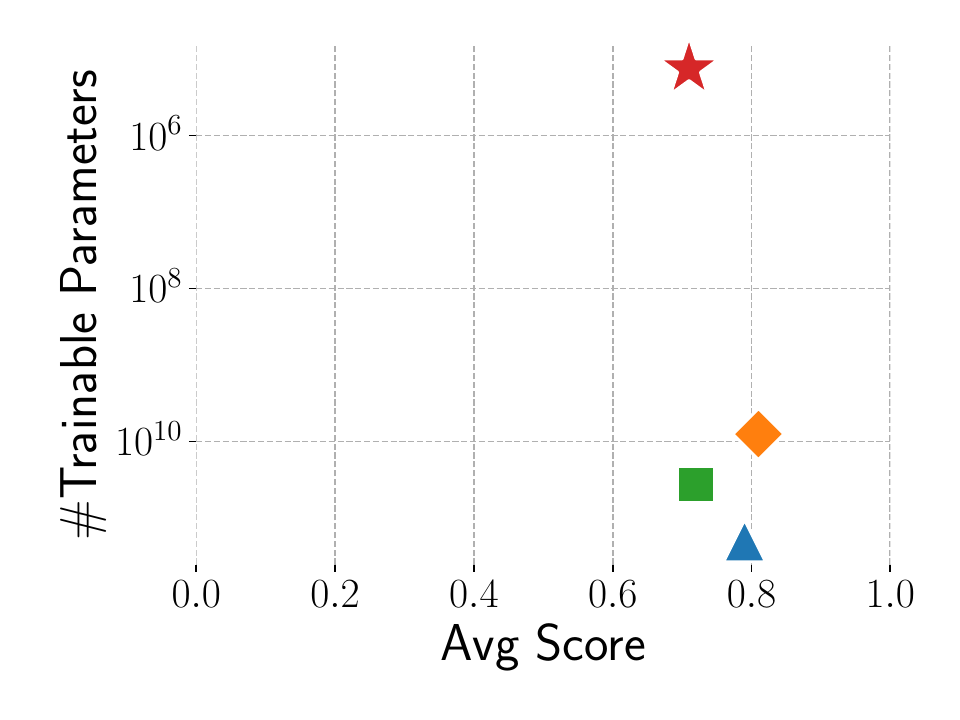}
    \caption{Accuracy-Efficiency Trade-off of our method compared to state-of-the-art LLMs. The X-axis represents the average score across all reasoning tasks, while the Y-axis reports the number of trainable parameters.}
    \label{fig:trade-off}
\end{figure}

\section{Results}
\label{sec:results}
\paragraph{Accuracy vs Efficiency} 
To address \textbf{RQ1}, we begin by validating our proposed method against the aforesaid established baselines. The results, summarized in Table~\ref{tab:unique-results}, illustrate the performance across the considered datasets. In the table, the best-performing scores are highlighted in bold, while the percentage improvements are calculated relative to the second-best results.

Overall, our method consistently outperforms the competing approaches across all the examined reasoning tasks, with substantial improvement in most configurations. The highest accuracy is achieved on the existence task, which represents the simplest form of reasoning. Conversely, the counting and identification tasks present greater challenges, and here the performance tends to fluctuate depending on the specific dataset, reflecting the increasing complexity required for these tasks.

In addition to comparing against classical baselines, we further evaluate our method against state-of-the-art LLMs to understand the trade-off between accuracy and computational efficiency. This comparison is visualized in Figure~\ref{fig:trade-off}, where the X-axis corresponds to the average score across all reasoning tasks, and the Y-axis indicates the number of trainable parameters. For this analysis, we select the synthetic dataset as a representative benchmark.

The figure reveals that while our method slightly underperforms relative to GPT-4o (0.79), o4-mini (0.81), and DeepSeek-R1 (0.72) — achieving an average score of 0.71 — it requires approximately $10^5$ fewer active parameters. This remarkable efficiency underscores the practicality of our approach, offering a highly competitive performance at a fraction of the computational cost, which is particularly advantageous for deployment in resource-constrained environments.

\begin{table*}[!ht]
        \centering
        
        \caption{Test accuracy over the considered datasets and reasoning tasks, varying the underlying KGE model.}
        \label{tab:unique-sensitivity}
        
        \resizebox{\linewidth}{!}{
        \begin{tabular}{w{c}{0.2cm}cccc|cccccc|ccccccccc}
        \toprule
        {} & {} & \multicolumn{3}{c}{\textbf{0-Hop}} & \multicolumn{6}{c}{\textbf{1-Hop}} & \multicolumn{9}{c}{\textbf{2-Hop}} \\
        \cmidrule(lr){3-5}
        \cmidrule(lr){6-11}
        \cmidrule(lr){12-20}
        {\textbf{Data}} & \textbf{Model} & \textbf{E-SL} & \textbf{C-SL} & \textbf{I-SL} & \textbf{E-SL} & \textbf{C-SL} & \textbf{I-SL} & \textbf{E-DL} & \textbf{C-DL} & \textbf{I-DL} & \textbf{E-SL} & \textbf{C-SL} & \textbf{I-SL} & \textbf{E-DL} & \textbf{C-DL} & \textbf{I-DL} & \textbf{E-TL} & \textbf{C-TL} & \textbf{I-TL} \\
        \midrule
        \multirow{4}{*}{\centering{\rotatebox[origin=c]{90}{\textbf{Synthetic}}}}  & TransE & 0.98 & 0.23 & 0.05 & 1.0 & 0.73 & 0.92 & 0.98 & 0.3 & 0.1 & 1.0 & 0.78 & 0.92 & 0.97 & 0.25 & 0.12 & 0.9 & 0.23 & 0.17 \\
 & DistMult & 1.0 & 0.63 & 0.08 & 1.0 & 0.83 & 0.77 & 1.0 & 0.17 & 0.1 & 1.0 & 0.83 & 0.77 & 0.95 & 0.67 & 0.08 & 0.9 & 0.6 & 0.18 \\
 & ComplEx & 1.0 & 0.7 & 0.08 & 1.0 & 0.68 & 0.78 & 1.0 & 0.63 & 0.08 & 1.0 & 0.68 & 0.78 & 0.95 & 0.67 & 0.13 & 0.9 & 0.67 & 0.18 \\
 & RotatE & 1.0 & 0.6 & 0.05 & 1.0 & 0.88 & 0.93 & 1.0 & 0.57 & 0.13 & 1.0 & 0.88 & 0.93 & 0.97 & 0.55 & 0.1 & 0.95 & 0.7 & 0.18 \\
\midrule
\multirow{4}{*}{\centering{\rotatebox[origin=c]{90}{\textbf{AIDS}}}}  & TransE & 1.0 & 0.29 & 0.06 & 1.0 & 0.21 & 0.18 & 0.88 & 0.46 & 0.41 & 0.96 & 0.23 & 0.18 & 0.9 & 0.26 & 0.1 & 0.92 & 0.5 & 0.46 \\
 & DistMult & 1.0 & 0.4 & 0.06 & 1.0 & 0.19 & 0.12 & 0.94 & 0.6 & 0.43 & 0.96 & 0.19 & 0.12 & 0.88 & 0.38 & 0.08 & 0.92 & 0.64 & 0.44 \\
 & ComplEx & 1.0 & 0.36 & 0.06 & 0.99 & 0.28 & 0.18 & 0.87 & 0.59 & 0.4 & 0.96 & 0.25 & 0.15 & 0.87 & 0.3 & 0.18 & 0.95 & 0.63 & 0.46 \\
 & RotatE & 1.0 & 0.28 & 0.06 & 1.0 & 0.21 & 0.14 & 0.95 & 0.62 & 0.41 & 0.97 & 0.21 & 0.14 & 0.88 & 0.39 & 0.1 & 0.95 & 0.57 & 0.46 \\
\midrule
\multirow{4}{*}{\centering{\rotatebox[origin=c]{90}{\textbf{MUTAG}}}}  & TransE & 1.0 & 0.37 & 0.37 & 1.0 & 0.58 & 0.16 & 0.89 & 0.84 & 0.68 & 1.0 & 0.42 & 0.05 & 0.89 & 0.58 & 0.58 & 1.0 & 0.74 & 0.63 \\
 & DistMult & 1.0 & 0.68 & 0.58 & 1.0 & 0.58 & 0.47 & 0.79 & 0.68 & 0.63 & 0.95 & 0.58 & 0.47 & 0.89 & 0.89 & 0.68 & 1.0 & 0.74 & 0.63 \\
 & ComplEx & 1.0 & 0.74 & 0.37 & 1.0 & 0.53 & 0.21 & 0.84 & 0.74 & 0.53 & 0.95 & 0.47 & 0.32 & 0.89 & 0.95 & 0.63 & 1.0 & 0.95 & 0.53 \\
 & RotatE & 1.0 & 0.74 & 0.32 & 1.0 & 0.53 & 0.11 & 0.79 & 0.84 & 0.53 & 1.0 & 0.53 & 0.11 & 0.79 & 0.79 & 0.58 & 1.0 & 0.84 & 0.63 \\
\midrule
\multirow{4}{*}{\centering{\rotatebox[origin=c]{90}{\textbf{AQSOL}}}}  & TransE & 1.0 & 0.19 & 0.05 & 1.0 & 0.21 & 0.19 & 0.9 & 0.41 & 0.36 & 0.85 & 0.22 & 0.19 & 0.93 & 0.27 & 0.11 & 0.96 & 0.43 & 0.36 \\
 & DistMult & 1.0 & 0.38 & 0.04 & 0.99 & 0.27 & 0.22 & 0.97 & 0.56 & 0.36 & 0.98 & 0.27 & 0.22 & 0.93 & 0.32 & 0.13 & 0.96 & 0.49 & 0.4 \\
 & ComplEx & 1.0 & 0.33 & 0.04 & 0.99 & 0.25 & 0.2 & 0.96 & 0.57 & 0.36 & 0.97 & 0.24 & 0.2 & 0.96 & 0.26 & 0.15 & 0.96 & 0.58 & 0.36 \\
 & RotatE & 1.0 & 0.38 & 0.06 & 0.99 & 0.21 & 0.21 & 0.96 & 0.52 & 0.4 & 0.97 & 0.21 & 0.21 & 0.91 & 0.28 & 0.17 & 0.96 & 0.52 & 0.41 \\
\bottomrule
        \end{tabular}
        }
    \end{table*}

\paragraph{Sensitivity Analysis} Further, we conduct a sensitivity analysis aimed at evaluating the robustness of our method when different KGE models are employed (\textbf{RQ2}). The full performance breakdown across all reasoning tasks is presented in Table~\ref{tab:unique-sensitivity}.

Our findings show no absolute winner across the considered embedders, but the performance varies depending on the dataset and the task at hand. Indeed, the results on the existence task remain relatively stable across different KGE models, indicating that simpler reasoning tasks are less sensitive to the choice of underlying representation. However, in the more demanding counting and identification tasks, the performance varies significantly depending on the selected embedder.

We hypothesize that this variability is due to several interacting factors: (i) the increased complexity inherent to the counting and identification tasks, (ii) the structural properties and difficulty levels embedded within the datasets, and (iii) the differing representational capacities of the various KGE models. These insights suggest that careful selection of the underlying embedding model is crucial when targeting more complex forms of reasoning.

\begin{figure}
    \centering
    \begin{minipage}{0.6\linewidth}
        \includegraphics[width=\linewidth]{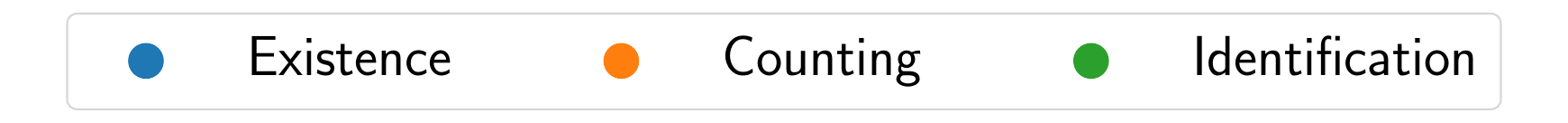}
    \end{minipage}
    \\
     \begin{subfigure}[b]{0.24\linewidth}
         \centering
         \includegraphics[width=\linewidth]{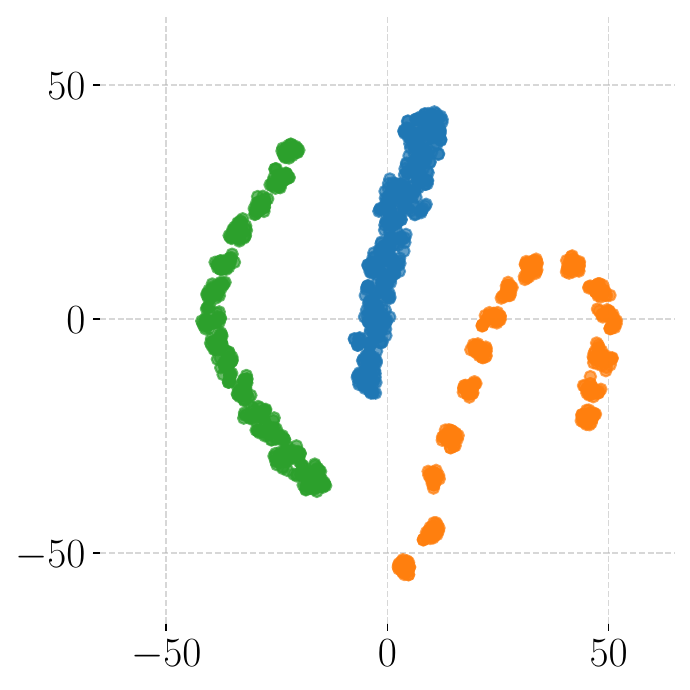}
         \caption{\textbf{TransE}}
     \end{subfigure}
     \hfill
     \begin{subfigure}[b]{0.24\linewidth}
         \centering
         \includegraphics[width=\linewidth]{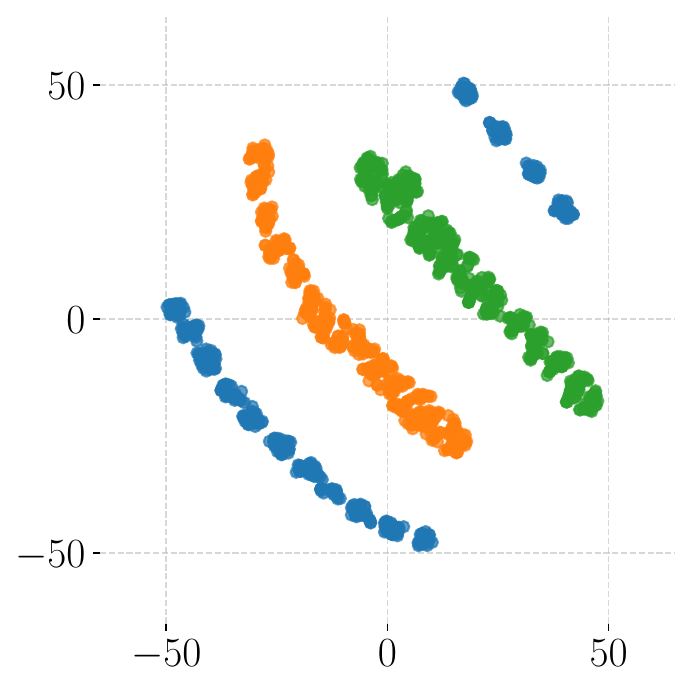}
         \caption{\textbf{DistMult}}
     \end{subfigure}
    \hfill
     \begin{subfigure}[b]{0.24\linewidth}
         \centering
         \includegraphics[width=\linewidth]{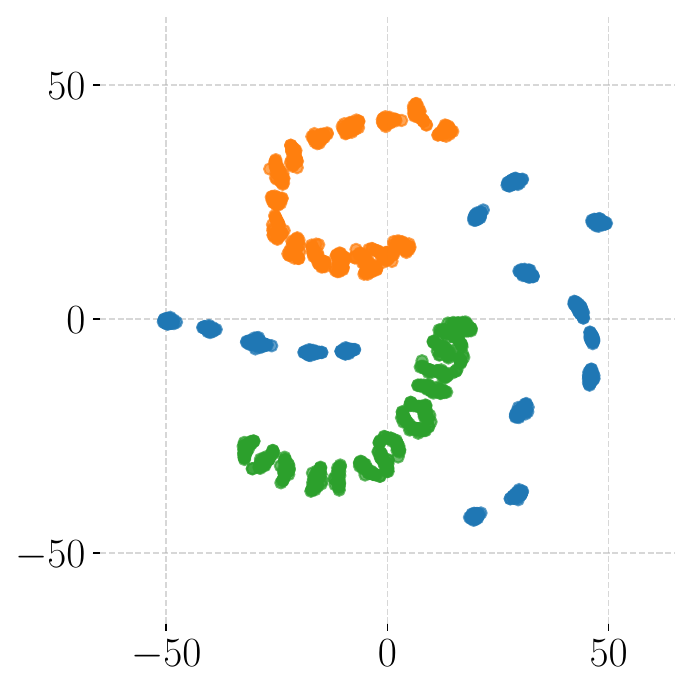}
         \caption{\textbf{ComplEx}}
     \end{subfigure}
     \hfill
     \begin{subfigure}[b]{0.24\linewidth}
         \centering
         \includegraphics[width=\linewidth]{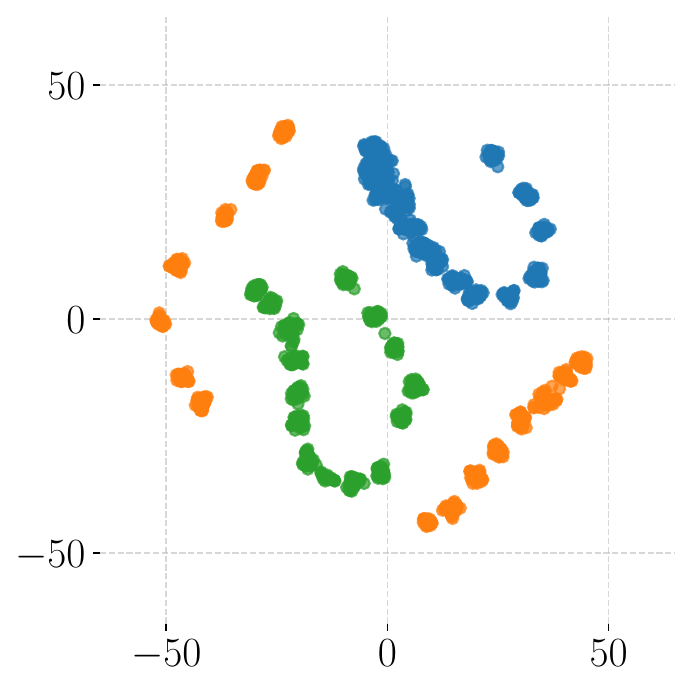}
         \caption{\textbf{RotatE}}
     \end{subfigure}
    \caption{t-SNE projection of the synthetic graph embeddings produced by each KGE model, trained on an Existence, Counting and Identification task.}
    \label{fig:embeddings}
\end{figure}

\paragraph{Latent Representation} Lastly, we address \textbf{RQ3} by examining how the final knowledge graph embeddings vary according to the specific reasoning task. To this end, we visualize the learned graph representations across different tasks. Figure~\ref{eq:kg-embedding} illustrates the t-SNE projections~\citep{tsne} of the embeddings obtained using each KGE model trained on an Existence, Counting, and Identification task, respectively.

This analysis reveals that: first, although the underlying graphs remain the \textbf{same} across tasks, their latent representations exhibit consistent adaptations depending on the reasoning objective; and secondly, such adaptations persist regardless of the underlying KGE model used. This behavior underscores the ability of our method not only to align the intrinsic representation of the knowledge graph with the latent space of the LLM, but also to flexibly encode task-relevant features, adjusting the structure of the embedding space to better support different types of reasoning.



\section{Conclusions and Future Work}
\label{sec:conclusions}

In this work, building upon prior research, we introduced a novel technique for the direct integration of Knowledge Graph (KG) representations into frozen Large Language Models (LLMs), eliminating the need for prompt engineering or costly LLM fine-tuning. Through extensive experimentation on both synthetic and real-world datasets, we evaluated our approach against a variety of baselines and state-of-the-art LLMs. The results consistently demonstrate the effectiveness, generalizability, and computational efficiency of our method across diverse reasoning tasks. Our findings validate the potential of integrating external knowledge representations to enhance LLMs' reasoning capabilities without sacrificing scalability or flexibility.

Despite these promising results, there remain several avenues for further improvement and extension. First, while our evaluation encompasses a wide range of reasoning tasks, it primarily focuses on node-related queries. Extending the framework to cover edge-centric and graph-level reasoning tasks would broaden its applicability and provide a more comprehensive validation of its generalization capabilities. Second, while our method shows robust performance across most tasks, certain complex reasoning scenarios, such as the Identification task, pose greater challenges. To address this, future work could explore the integration of more sophisticated learning techniques during training, such as task-specific inductive biases, or hybrid approaches combining symbolic and neural representations.
Finally, we envision that adapting the method to support dynamic or evolving knowledge graphs, as well as exploring tighter coupling mechanisms between the KG embeddings and the LLM latent space, could further enhance performance and unlock new capabilities for knowledge-intensive applications.


\bibliographystyle{unsrtnat}


\end{document}